\documentclass{article} 
\usepackage{iclr2021_conference,times}

\usepackage{amsmath,amsfonts,bm}

\def\eqref#1{equation~\ref{#1}}

\def\1{\bm{1}}

\DeclareMathAlphabet{\mathsfit}{\encodingdefault}{\sfdefault}{m}{sl}
\SetMathAlphabet{\mathsfit}{bold}{\encodingdefault}{\sfdefault}{bx}{n}

\usepackage{hyperref}
\usepackage{url}

\usepackage{graphicx}
\usepackage{subcaption}
\usepackage{multirow}
\usepackage{multicol}
\usepackage{booktabs}

\title{Document AI: Benchmarks, Models and Applications}

\author{Lei Cui, Yiheng Xu, Tengchao Lv, Furu Wei\\
Microsoft Research Asia\\
\texttt{\{lecu,t-yihengxu,tengchaolv,fuwei\}@microsoft.com} \\
}

\iclrfinalcopy 
\begin{document}

\maketitle

\begin{abstract}
Document AI, or Document Intelligence, is a relatively new research topic that refers to the techniques for automatically reading, understanding, and analyzing business documents. It is an important research direction for natural language processing and computer vision. In recent years, the popularity of deep learning technology has greatly advanced the development of Document AI, such as document layout analysis, visual information extraction, document visual question answering, document image classification, etc. This paper briefly reviews some of the representative models, tasks, and benchmark datasets. Furthermore, we also introduce early-stage heuristic rule-based document analysis, statistical machine learning algorithms, and deep learning approaches especially pre-training methods. Finally, we look into future directions for Document AI research.
\end{abstract}

\section{Document AI}

Document AI, or Document Intelligence, is a booming research topic with increased industrial demand in recent years. It mainly refers to the process of automated understanding, classifying and extracting information with rich typesetting formats from webpages, digital-born documents or scanned documents through AI technology. Due to the diversity of layouts and formats, low-quality scanned document images, and the complexity of the template structure, Document AI is a very challenging task and has attracted widespread attention in related research areas. With the acceleration of digitization, the structured analysis and content extraction of documents, images and others has become a key part of the success of digital transformation. Meanwhile automatic, accurate, and rapid information processing is crucial to improving productivity. Taking business documents as an example, they not only contain the processing details and knowledge accumulation of a company's internal and external affairs, but also a large number of industry-related entities and digital information. Manually extracting  information is time-consuming and labor-intensive with low accuracy and low reusability. Document AI deeply combines artificial intelligence and human intelligence, and has different types of applications in multiple industries such as finance, healthcare, insurance, energy and logistics. For instance, in the finance field, it can conduct financial report analysis and intelligent decision analysis, and provide scientific and systematic data support for the formulation of corporate strategies and investment decisions. In healthcare, it can improve the digitization of medical cases and enhance diagnosis accuracy. By analyzing the correlation between medical literature and cases, people can locate potential treatment options. In the accounting field, it can achieve automatic information extraction of invoices and purchase orders, automatically analyze a large number of unstructured documents, and support different downstream business scenarios, saving a lot of manual processing time.

\begin{figure*}[ht]
    \centering
    \includegraphics[width=0.9\textwidth]{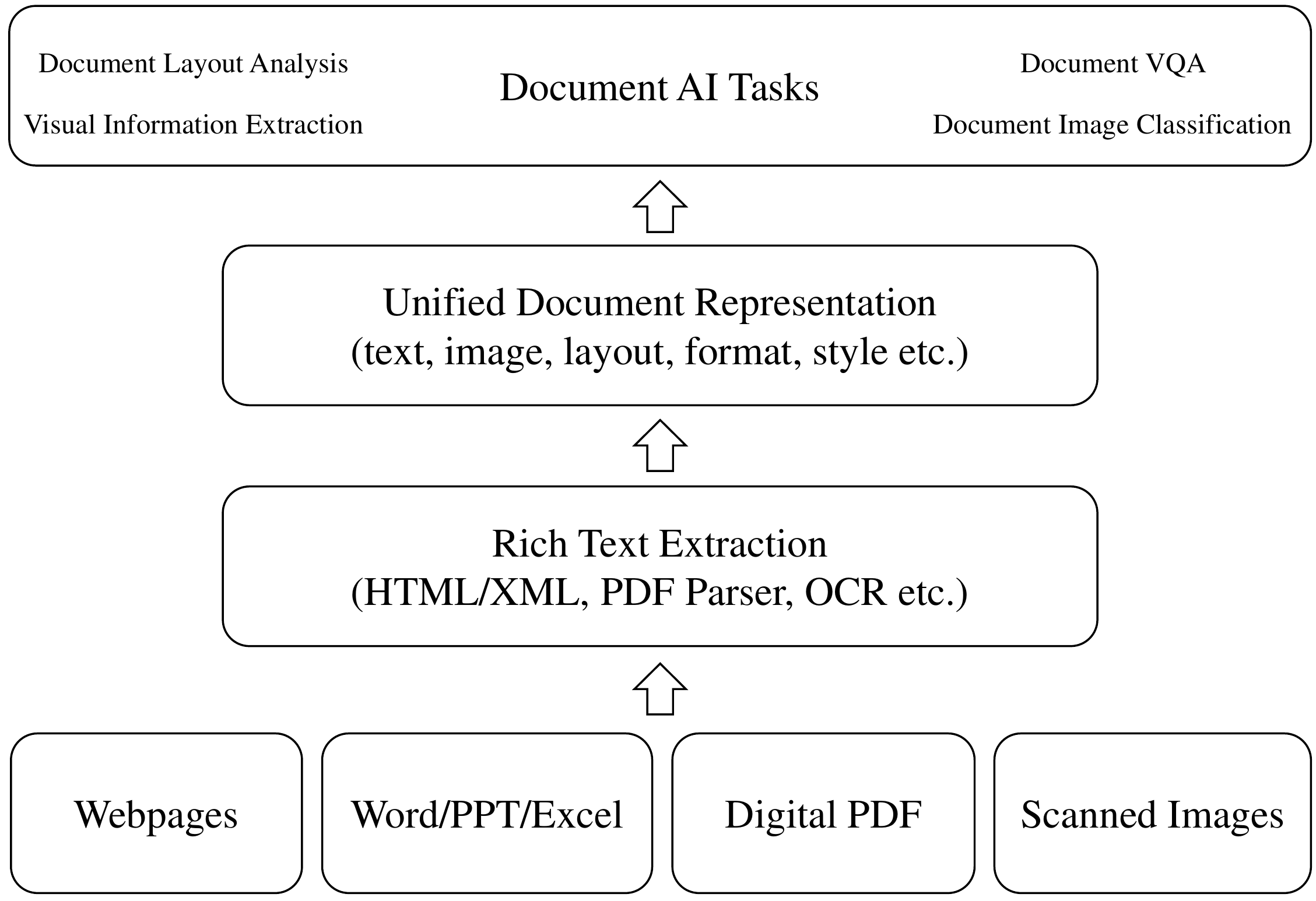}
    \caption{Overview of Document AI}
    \label{fig:1}
\end{figure*}

Over the past few decades, the development of document intelligence has roughly gone through different stages, evolving from simple rule-based heuristics to neural network approaches. In the early 1990s, researchers mostly used rule-based heuristic approaches for document understanding and analysis. By manually observing the layout information of documents, they summarized some heuristic rules and processed documents with fixed layout information. However, traditional rule-based methods often require large labor costs, and these manually summarized rules are not scalable. Therefore, researchers have begun to adopt methods based on statistical machine learning. Since 2000, with the development of machine learning technology, machine learning models based on annotated data have become the mainstream method of document processing. It uses artificially designed feature templates to learn the weights of different features to understand and analyze the content and layout of a document. Although annotated data is leveraged for supervised learning and previous methods can bring a certain degree of performance improvement, the general usability is often not satisfactory due to the lack of customized rules and the number of training samples. Additionally, the migration and adaptation costs for different types of documents are relatively high, making previous approaches not practical for widespread commercial use. In recent years, with the development of deep learning technology and the accumulation of a large number of unlabeled electronic documents, document analysis and recognition technology has entered a new era. Figure~\ref{fig:1} represents the basic framework of Document AI technology under the current deep learning framework, in which different types of documents are extracted through content extraction tools (HTML/XML extraction, PDF parser, OCR, etc.) where text content, layout information, and visual image information are well organized. Then, large-scale deep neural networks are pre-trained and fine-tuned to complete a variety of downstream Document AI tasks, including document layout analysis, visual information extraction, document visual question answering, and document image classification etc. The emergence of deep learning, especially of the pre-training technique represented by Convolutional Neural Networks (CNN), Graph Neural Networks (GNN) and Transformer architecture~\citep{vaswani2017attention}, has completely shifted the traditional machine learning paradigm that requires a lot of manual annotations. Instead, it heavily relies on a large amount of unlabeled data for self-supervised learning, and addresses the downstream tasks through the "pre-training and fine-tuning" paradigm which leads to a significant breakthrough in Document AI tasks. We have also observed many successful Document AI products, such as Microsoft Form Recognizer\footnote{\url{https://azure.microsoft.com/en-us/services/form-recognizer/}}, Amazon Textract\footnote{\url{https://aws.amazon.com/textract}}, Google Document AI\footnote{\url{https://cloud.google.com/document-ai}} and many others, which have fundamentally empowered a variety of industries with the Document AI technology.

Although deep learning has greatly improved the accuracy of Document AI tasks, there are still many problems to be solved in practical applications. First, due to the limitation of the input length of current large-scale pre-trained language models, they usually need to truncate documents into several parts to be input to the model for processing, which poses a great challenge for the multi-page and cross-page understanding of complex long documents. Second, due to the quality mismatch between annotated training data and document images in real-world business which usually comes from the scanning equipment, crumpled paper and random placement, poor performance is observed and more data synthesis/augmentation techniques are needed to help existing models improve the performance. Third, the current Document AI tasks are often trained independently, and the correlation between different tasks has not been effectively leveraged. For instance, visual information extraction and document visual question answering have some common semantic representations, which can be better solved by using a multi-task learning framework. Finally, the pre-trained Document AI models also encountered the problem of insufficient computing resources and labeled training samples in practical applications. Therefore, model compression, few-shot learning and zero-shot learning are important research directions at present and have great practical values.

Next, we introduce the current mainstream Document AI models (CNN, GNN and Transformer), tasks, and benchmark datasets, and then elaborate on early-stage document analysis techniques based on heuristic rules, algorithms, and models based on traditional statistical machine learning, as well as the most recent deep learning models, especially the multimodal pre-training technique. Finally, we outline future directions of Document AI research.

\begin{figure*}[t]
    \centering
    \includegraphics[width=0.8\textwidth]{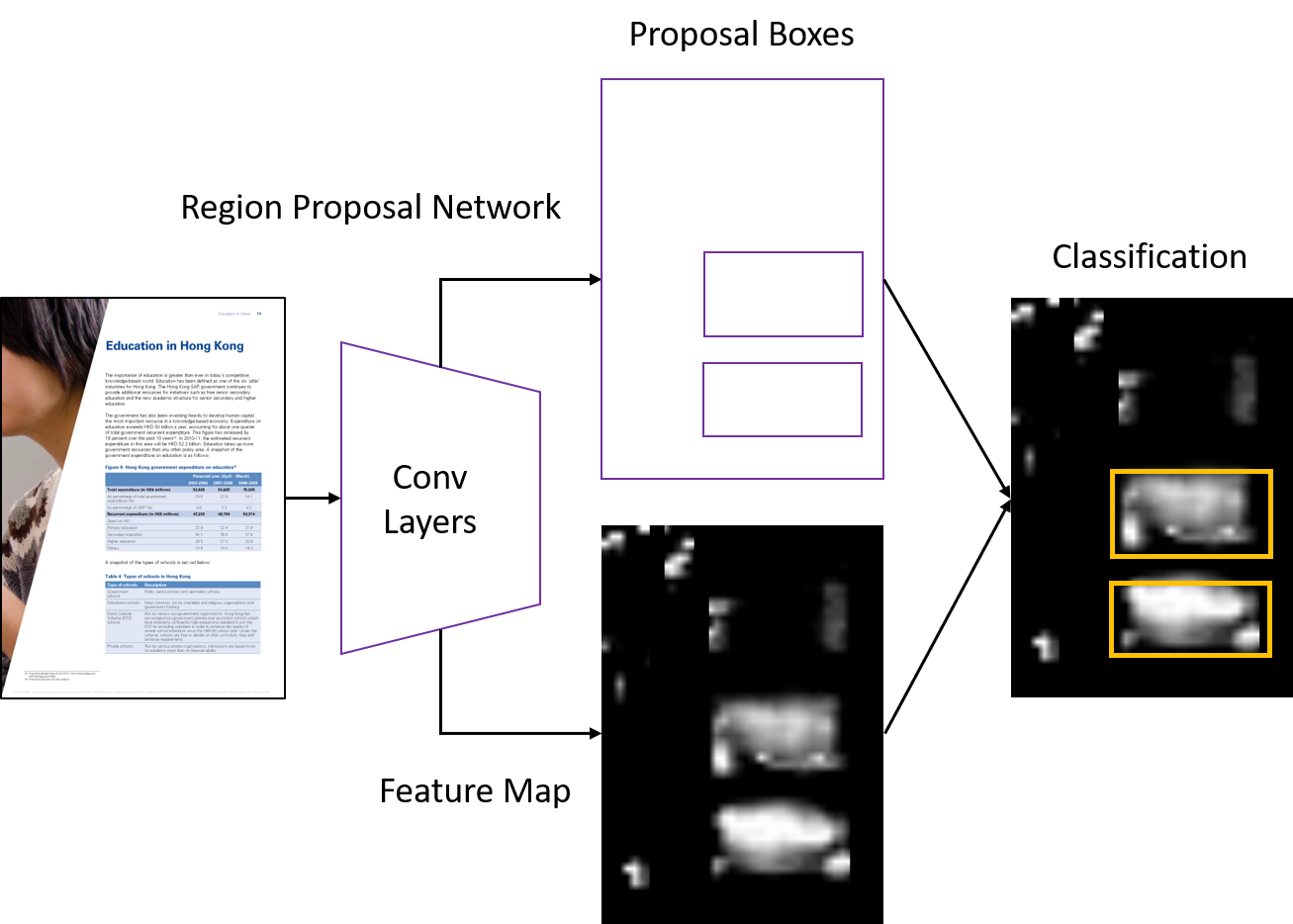}
    \caption{Document layout analysis with Faster R-CNN}
    \label{fig:2}
\end{figure*}

\section{Representative Models, Tasks and Benchmarks}

\subsection{Document Layout Analysis with Convolutional Neural Networks}
\label{sectioncnn}
In recent years, convolutional neural networks have achieved great success in the field of computer vision, especially the supervised pre-training model ResNet~\citep{he2015deep} based on large-scale annotated datasets ImageNet and COCO has brought great performance improvements in image classification, object detection and scene segmentation. Specifically, with multi-stage models such as Faster R-CNN~\citep{ren2016faster} and Mask R-CNN~\citep{he2018mask}, and single-stage detection models including SSD~\citep{Liu_2016} and YOLO~\citep{yolov3}, object detection has almost become a solved problem in computer vision. Document layout analysis can essentially be regarded as an object detection task for document images. Basic units such as headings, paragraphs, tables, figures and charts in the document are the objects that need to be detected and recognized. \cite{yang2017learning} regard document layout analysis as a pixel-level segmentation task, and used convolutional neural networks for pixel classification to achieve good results. \cite{8270123} first apply the Faster R-CNN model to table detection and recognition in document layout analysis as shown in Figure~\ref{fig:2}, achieving SOTA performance in the ICDAR 2013 table detection dataset~\citep{Gbel2013ICDAR2T}. Although document layout analysis is a classic document intelligence task, it has been limited to a small training dataset for many years, which is not sufficient for applying  pre-trained models in computer vision. With large-scale weakly supervised document layout analysis datasets such as PubLayNet~\citep{zhong2019publaynet}, PubTabNet~\citep{zhong2019image}, TableBank~\citep{li-etal-2020-tablebank} and DocBank~\citep{li-etal-2020-docbank}, researchers can conduct a more in-depth comparison and analysis of different computer vision models and algorithms, and further promote the development of document layout analysis techniques.

\begin{figure*}[t]
\centering
    \begin{subfigure}[b]{1\textwidth}
        \includegraphics[width=\textwidth]{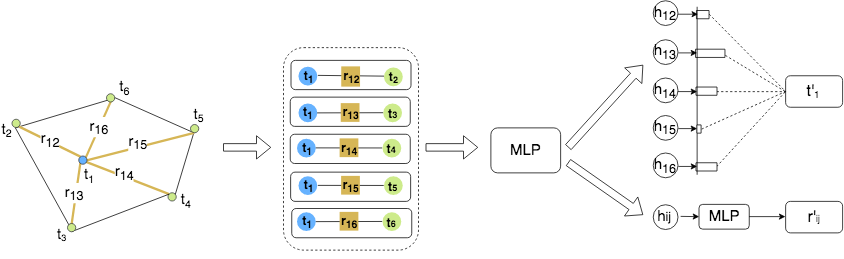}
        \caption{}
        \label{fig:3a}
    \end{subfigure}
    \begin{subfigure}[b]{0.5\textwidth}
        \includegraphics[width=\textwidth]{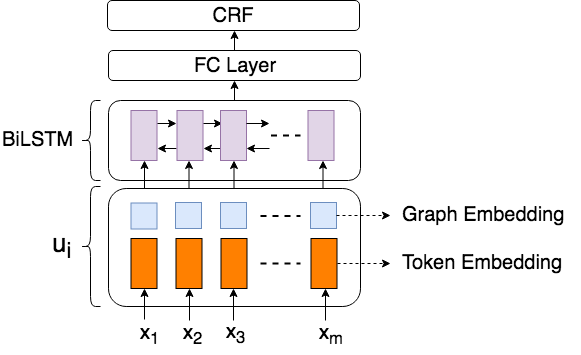}
        \caption{}
        \label{fig:3b}
    \end{subfigure}
    \caption{Visual information extraction with GNN}\label{fig:3}
\end{figure*}

\subsection{Visual Information Extraction with Graph Neural Networks}
\label{sectiongnn}
Information extraction is the process of extracting structured information from unstructured text, and it has been widely studied as a classical and fundamental NLP problem. Traditional information extraction focuses on extracting entity and relationship information from plain text, but less research has been done on visually rich documents. Visually rich documents refers to the text data whose semantic structure is not only determined by the content of the text, but also related to visual elements such as layout, typesetting formats as well as table/figure structures. Visually-rich documents can be found everywhere in real-world applications, such as receipts, certificates, insurance files, etc. \cite{liu-etal-2019-graph} propose  modeling visually rich documents using graph convolutional neural networks. As shown in Figure~\ref{fig:3}, each image is passed through the OCR system to obtain a set of text blocks, each of which contains information about its coordinates in the image with the text content. This work constitutes this set of text blocks as a fully connected directed graph, i.e., each text block constitutes a node, and each node is connected to all other nodes. The initial features of the nodes are obtained from the text content of the text blocks by Bi-LSTM encoding. The initial features of the edges are the relative distance between the neighboring text blocks and the current text block and the aspect ratio of these two text blocks. Unlike other graph convolution models that only convolve on nodes, this work focuses more on the "individual-relationship-individual" ternary feature set in information extraction, so convolution is performed on the "node-edge-node" ternary feature set. In addition, the self-attention mechanism allows the network to select more noteworthy information in all directed triads in fully connected directed graphs and aggregate the weighted features. The initial node features and edge features are convolved in multiple layers to obtain the high-level representations of nodes and edges. Experiments show that this graph convolution model significantly outperforms the Bi-LSTM+CRF models.  In addition, experiments have shown that visual information plays a major role, increasing the discrimination of texts with similar semantics. Text information also plays a certain auxiliary role to visual information. The self-attention mechanism is basically not helpful for fixed layout data, but it generates some level of improvement on non-fixed layout data.

\subsection{General-purpose Multimodal Pre-training with the Transformer Architecture}
\label{sectiontransformer}

In many cases, the spatial relationship of text blocks in a document usually contains rich semantic information. For instance, forms are usually displayed in the form of key-value pairs. Typically, the arrangement of key-value pairs is usually in the left-right order or the up-down order. Similarly, in a tabular document, the text blocks are usually arranged in a grid layout and the header usually appears in the first column or row. This layout invariance among different document types is a critical property for general-purpose pre-training. Through pre-training, the position information that is naturally aligned with the text can provide richer semantic information for downstream tasks. For visually-rich documents, in addition to positional information, the visual information presented with the text can also help downstream tasks, such as font types, sizes, styles and other visually-rich formats. For instance, in forms, the key part of a key-value pair is usually given in bold form. In general documents, the title of the article will usually be enlarged and bold, and the nouns of special concepts will be displayed in italics, etc. For document-level tasks, the overall visual signals can provide global structural information, and there is a clear visual difference between different document types, such as a personal resume and a scientific paper. The visual features displayed in these visually-rich documents can be extracted by visual encoders and combined into the pre-training stage, thereby effectively improving downstream tasks.

\begin{figure*}[t]
    \centering
    \includegraphics[width=1\textwidth]{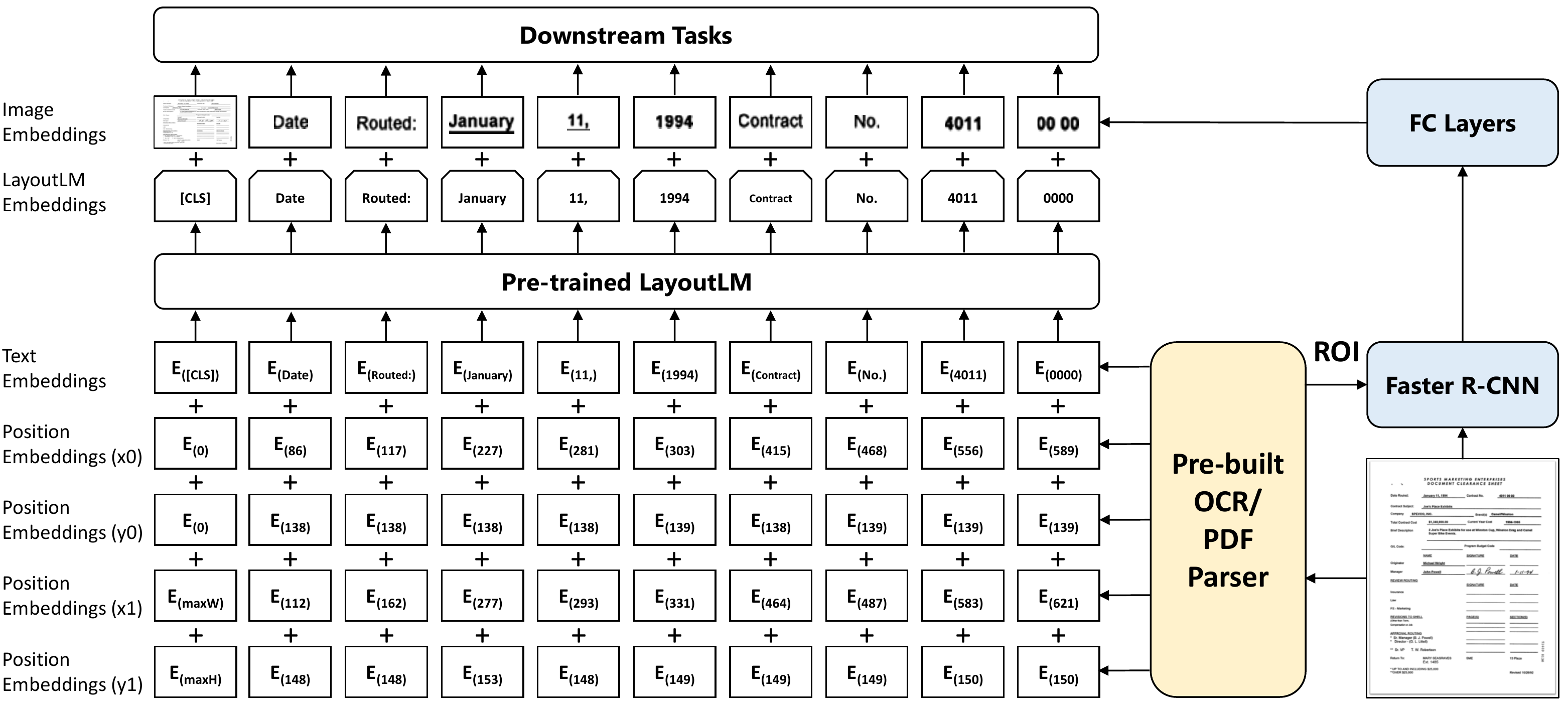}
    \caption{The LayoutLM architecture, where 2-D layout and image embeddings are integrated into the Transformer architecture.}
    \label{fig:4}
\end{figure*}

To leverage the layout and visual information, \cite{10.1145/3394486.3403172} propose a general document pre-training model LayoutLM~\citep{10.1145/3394486.3403172}, as shown in Figure~\ref{fig:4}. Two new embedding layers, 2-D position embedding and image embedding are added on the basis of the existing pre-trained model, so that the document structure and visual information can be effectively combined. Specifically, according to the text bounding boxes obtained by OCR, the algorithm first gets the coordinates of the text in the document. After converting the corresponding coordinates into virtual coordinates, the model calculates the representation of the coordinates corresponding to the four embedding sub-layers of x, y, w, and h. The final 2-D position embedding is the sum of the embedding of the four sub-layers. In image embedding, the model considers the bounding boxes corresponding to each text as the proposal in the Faster R-CNN to extract the corresponding local features. In particular, since the [CLS] symbol is used to represent the semantics of the entire document, the model also uses the entire document image as the image embedding at this position to maintain multimodal alignment.

In the pre-training stage, the authors propose two self-supervised pre-training tasks for LayoutLM:
\paragraph{Task \#1: Masked Visual-Language Model}
Inspired by the masked language model, the authors propose the Masked Visual-language Model (MVLM) to learn language representation with the clues of 2-D position embeddings and text embeddings. During pre-training, the model randomly masks some of the input tokens but keeps the 2-D position embeddings and other text embeddings. The model is then trained to predict the masked tokens given the context. In this way, the LayoutLM model not only understands the language contexts, but also utilizes the corresponding 2-D position information, thereby bridging the gap between the visual and language modalities.

\paragraph{Task \#2: Multi-Label Document Classification}
For document image understanding, many tasks require the model to generate high-quality document-level representations. As the IIT-CDIP Test Collection includes multiple tags for each document image, the model also uses a Multi-label Document Classification (MDC) loss during the pre-training phase. Given a set of scanned documents, the model uses the document tags to supervise the pre-training process so that the model can cluster the knowledge from different domains and generate better document-level representation. Since the MDC loss needs the label for each document image that may not exist for larger datasets, it is optional during the pre-training and may not be used for pre-training larger models in the future.

Experiments show that pre-training with layout and visual information can be effectively transferred to downstream tasks. Significant accuracy improvements are achieved in multiple downstream tasks. Different from the convolutional neural networks and graph neural networks, the advantage of the general document-level pre-training model is that it can support different types of downstream applications.

\begin{table}[t]
\begin{center}
\begin{tabular}{|c|l|l|l|}
\hline  \bf Task  & \bf Benchmark & \bf Langauge &\bf Paper/Link \\ \hline
\multirow{14}{*}{Document Layout Analysis} 

& ICDAR 2013  & En & \cite{Gbel2013ICDAR2T} \\ \cline{2-4}
& ICDAR 2019 & En & \cite{8978120} \\\cline{2-4}
& ICDAR 2021 & En & \cite{yepes2021icdar} \\\cline{2-4}
& UNLV & En & \cite{10.1145/1815330.1815345} \\\cline{2-4}
& Marmot & Zh/En & \cite{fang2012dataset} \\\cline{2-4}
& PubTabNet & En & \cite{zhong2019image} \\\cline{2-4}
& PubLayNet & En & \cite{zhong2019publaynet} \\\cline{2-4}
& TableBank & En & \cite{li-etal-2020-tablebank} \\\cline{2-4}
& DocBank & En & \cite{li-etal-2020-docbank} \\\cline{2-4}
& TNCR & En & \cite{abdallah2021tncr} \\\cline{2-4}
& TabLeX & En &  \cite{desai2021tablex} \\\cline{2-4}
& PubTables & En & \cite{smock2021pubtables1m}\\\cline{2-4}
& IIIT-AR-13K & En & \cite{mondal2020iiit} \\\cline{2-4}
& ReadingBank & En & \cite{wang2021layoutreader} \\
\hline
\multirow{10}{*}{Visual Information Extraction} 
& SWDE & En & \cite{10.1145/2009916.2010020} \\\cline{2-4}
& FUNSD & En & \cite{jaume2019} \\\cline{2-4}
& SROIE & En & \cite{Huang_2019}  \\\cline{2-4}
& CORD & En & \cite{park2019cord} \\\cline{2-4}
& EATEN & Zh & \cite{guo2019eaten} \\\cline{2-4}
& EPHOIE & Zh & \cite{wang2021towards} \\\cline{2-4}
& Deepform & En & \cite{deepform} \\\cline{2-4}
& Kleister & En & \cite{stanislawek2021kleister} \\\cline{2-4}
& XFUND & \begin{tabular}{@{}l@{}}Zh/Ja/Es/\\Fr/It/De/Pt  \end{tabular}& \cite{xu2021layoutxlm} \\
\hline
\multirow{5}{*}{Document VQA}
& DocVQA & En & \cite{mathew2021docvqa} \\\cline{2-4}
& InfographicsVQA & En & \cite{mathew2021infographicvqa} \\\cline{2-4}
& VisualMRC & En & \cite{tanaka2021visualmrc} \\\cline{2-4}
& WebSRC & En & \cite{chen2021websrc} \\\cline{2-4}
& Insurance VQA & Zh & \url{https://bit.ly/36O2Vow} \\
\hline
\multirow{2}{*}{Document Image Classification}
& Tobacco-3482 & En& \cite{Kumar2014StructuralSF} \\\cline{2-4}
& RVL-CDIP & En & \cite{harley2015icdar} \\
\hline
\end{tabular}
\end{center}
\caption{Benchmark datasets for document layout analysis, visual information extraction, document visual question answering and document image classification.}
\label{tab:1}
\end{table}

\subsection{Mainstream Document AI Tasks and Benchmarks}

Document AI involves automatic reading, comprehension, and analysis of documents. In real-world application scenarios, it mainly includes four types of tasks, namely:
\paragraph{Document Layout Analysis}
This is the process of automatic analysis, recognition and understanding of images, text, table/figure/chart information and positional relationships in the document layout.

\paragraph{Visual Information Extraction}
This refers to the techniques of extracting entities and their relationships from a large amount of unstructured content in a document. Unlike traditional pure text information extraction, the construction of the document turns the text from a one-dimensional sequential arrangement into a two-dimensional spatial arrangement. This makes text information, visual information and layout information extremely important influencing factors in visual information extraction.

\paragraph{Document Visual Question Answering}
Given digital-born documents or scanned images, PDF parsing, OCR or other text extraction tools are first used to automatically recognize the textual content, the system needs to answer the natural language questions about the documents by judging the internal logic of the recognized text.

\paragraph{Document Image Classification}
This refers to the process of analyzing and identifying document images, while classifying them into different categories such as scientific papers, resumes, invoices, receipts and many others.

For these four main Document AI tasks, there have been a large number of open-sourced benchmark datasets in academia and industry, as shown in the Table~\ref{tab:1}. This has greatly promoted the construction of new algorithms and models by researchers in related research areas, especially the most recent deep learning based models that achieve SOTA performance in these tasks. Next, we will introduce in detail the classic models and algorithms in different periods in the past, including document analysis techniques based on heuristic rules, document analysis technology based on statistical machine learning, and general Document AI models based on deep learning.

\section{Heuristic Rule-based Document Layout Analysis}

Document layout analysis using heuristic rules can be roughly divided into three ways: top-down, bottom-up, and hybrid strategy. The top-down methods divide a document image into different areas step by step. Cutting is performed recursively until the area is divided to a predefined standard, usually blocks or columns. The bottom-up methods use pixels or components as the basic element units, where the basic elements are grouped and merged to form a larger homogeneous area. The top-down approach enables faster and more efficient analysis of documents in specific formats, while bottom-up approaches require more computation resources but are more versatile and can cover more documents with different layout types. The hybrid strategy combines the top-down and bottom-up to produce better results.

This section introduces document analysis techniques from the top-down and bottom-up perspectives, including projection profile, image smearing, connected components and others.

\subsection{Projection Profile}

ccProjection profile is widely used in document analysis as a top-down analysis method. \cite{nagy1984hierarchical} use the X-Y cut algorithm to cut the document. This method is suitable for structured text with fixed text areas and line spacing, but it is sensitive to boundary noise and cannot provide good results on slanted text. ~\cite{bar2009line} use the dynamic local projection-profile to calculate the inclination of the document in an attempt to eliminate  performance degradation caused by text skew. Experiments have proven that the model has obtained more accurate results on slanted and curved text. In addition, many variations of the X-Y cut algorithm have been proposed to address existing problems in document analysis. \cite{o1993document} extends the X-Y cut algorithm to use the projection of the component bounding boxes, and \cite{sylwester1995trainable} use an evaluation metric called edit-cost to guide the segmentation model, which improves overall performance.

The projection profile analysis is suitable for structured text, especially documents with a manhattan-based layout. The performance may not be satisfactory for documents with complex layouts, slanted text, or border noises.

\subsection{Image Smearing}

Image smearing refers to permeating from one location to the surroundings and gradually expanding to all homogeneous areas to determine its layout in the page. \cite{wong1982document} adopt a top-down strategy and uses Run-Length Smoothing Algorithm (RLSA) to determine homogeneous regions. After the image is binarized, the pixel value 0 represents the background, and 1 is the foreground. When the number of 0s around 0 is less than the specified threshold $C$, the 0 at that position is changed to 1, and the RLSA uses this operation to merge the foreground that is nearby into a whole unit. In this way, characters can be gradually merged into words, and words can be merged into lines of text, and then the range continues to extend to the entire homogeneous area. On this basis, \cite{fisher1990rule} go further by adding preprocessing such as noise removal and tilt correction. In addition, the threshold $C$ of RLSA is modified according to the dynamic algorithm to further improve adaptability. \cite{esposito1990experimental} use a similar approach but the operation object is changed from pixels to character frames. \cite{shi2004line} expand each position pixel in the image to obtain a new grayscale image, which is extracted and shows good performance in the case of handwritten fonts, text slanting, etc.

\subsection{Connected Components}

As a bottom-up approach, connected components infers the relationship among the small elements, which is used to find homogeneous regions and finally classifies the regions into different layouts. \cite{fisher1990rule} use connected components to find the K-Nearest Neighbors (KNN) components of each component and infer the attributes of the current area through the relationship between the positions and angles of each other. \cite{saitoh1993document} merge the text into lines according to the inclination of the document, and then merge the lines into regions and classify them into different attributes. \cite{kise1998segmentation} also try to solve the problem of text skew. The authors use an approximated area Voronoi diagram to obtain the candidate boundary of the area. This operation is effective for areas with any angle of inclination. However, due to the need to estimate character spacing and line spacing during the calculation process, the model cannot perform well when the document contains large fonts and wide character spacing. In addition, \cite{bukhari2010document} also use AutoMLP on the basis of connected components in order to find the best parameters of the classifier to further improve the performance.

\subsection{Other Approaches}

In addition to the above methods, there are some other heuristic rule based document layout analysis approaches. \cite{baird1990image} use a top-down approach to divide the document into areas by blanks. \cite{xiao2003text} use the Delaunay Triangulation algorithm for document analysis. On this basis, \cite{bukhari2009script} apply it to script-independent handwritten documents. In addition, there are some hybrid models. \cite{okamoto1993hybrid} use separators and blanks to cut blocks, and further merge internal components into text lines in each block. \cite{smith2009hybrid} divide the document analysis into two parts. First, the bottom-up method is used to locate the tab characters, and the column layout is inferred with the help of the tab characters. Second, it uses a top-down approach to infer the structure and text order on the column layout.

\section{Machine Learning based Document Layout Analysis}

The machine learning based document analysis process is usually divided into two stages: 1) segmenting the document image to obtain multiple candidate regions; 2) classifying the document regions and distinguishing them such as text blocks and images. Some research work tries to use machine learning algorithms for document segmentation, while the rest try to construct features on generated regions and use machine learning algorithms to classify the regions. In addition, due to the performance boost led by machine learning, more machine learning models have been tried in table detection tasks, since table detection is a vital subtask of document analysis. This section will introduce machine learning approaches for different layout analysis tasks.

\subsection{Document Segmentation}

For document segmentation, \cite{baechler2011multi} combine the X-Y cut algorithm and use logistic regression to segment the document and discard the blank areas. After obtaining the corresponding regions, they also compare the performance of algorithms such as KNN, logistic regression and Maximum Entropy Markov Models (MEMM) as classifiers. The experiment shows that MEMM and logistic regression have better performance on classification tasks. \cite{esposito2008machine} further strengthen machine learning algorithms in document segmentation. In a bottom-up way, a kernel-based algorithm~\citep{dietterich1997solving} is used in the process of merging letters to words and text lines, and the results are converted into an xml structure for storage. After that, the Document Organization Composer (DOC) algorithm is used to analyze the documents. \cite{wu2008machine} focus on the problem of two reading orders of text at the same time. The existing models assume that the text information has only one reading order, but it cannot work normally when it encounters texts written in horizontal or vertical directions, such as in Chinese or Japanese. The proposed model divides the document segmentation process into four steps for judging and processing the text, and used the Support Vector Machines (SVM) model to decide whether to execute these steps in a pre-defined order.

\subsection{Region Classification}

For region classification, conventional research work usually leverages machine learning models to distinguish different regions. \cite{wei2013evaluation} compare the advantages and disadvantages of SVM, multi-layer perceptrons (MLP) and Gaussian Mixture Models (GMM) as classifiers. Experiments show that the classification accuracy of SVM and MLP are significantly better than GMM in region classification. \cite{bukhari2012layout} manually construct and extract multiple features from the document regions, and then used the AutoMLP algorithm to classify them. The classification accuracy of 95\% is obtained in the Arabic dataset. \cite{baechler2011multi} further improve the performance in region classification using a pyramid algorithm by conducting three levels of analysis on medieval manuscripts and using Dynamic Multi-Layer Perceptron (DMLP) as the classification model.

\subsection{Table Detection}

In addition to the above methods, there is a lot of research using traditional machine learning models for table detection and recognition. \citep{wang2000improvement,wangt2001automatic,wang2002table} use a binary tree to analyze the document in a top-down way to find the candidate table areas, and determine the final table area according to the predefined features. \cite{pinto2003table} use a Conditional Random Field (CRF) model to extract the table area in the HTML page, and identify the title, subtitle and other content in the table. \cite{e2009learning} uses Hidden Markov Models (HMM) to extract table regions. \cite{chen2011table} retrieve the table area in the handwritten document and use an SVM model to identify the texts within that region, and predict the location of the table based on the text lines. \cite{kasar2013learning} identify the horizontal and vertical lines in the figure, and then use an SVM model to classify the attributes of each line to determine whether the line belongs to the table. \cite{barlas2014typed} use an MLP model to classify the connected component in the document and determined whether it belongs to tables. \cite{bansal2014table} use the leptonica library~\cite{bloomberg1991multiresolution} to segment the document, and then construct features containing surrounding information for each region. By using the fix-point model~\cite{li2013fixed} to identify the table areas, the model does not only conduct the region classification, but also learns the relationship among different areas. \cite{rashid2017table} take the region classification into the word level and then used AutoMLP to determine whether the word belongs to the table.

\section{Deep Learning based Document AI}

In recent years, deep learning methods have become a new paradigm for solving many machine learning problems. Deep learning methods have been confirmed to be effective in many research areas. Recently, the popularity of pre-trained models has further improved the performance of deep neural networks. The development of Document AI also reflects a similar trend with other applications in deep learning. In this section, we divide the existing models into two parts: deep learning models for specific tasks and general-purpose pre-trained models that support a variety of downstream tasks.

\subsection{Task-specific Deep Learning Models}

\subsubsection{Document Layout Analysis}

Document layout analysis includes two main subtasks: visual analysis and semantic analysis~\citep{binmakhashen2019document}. The main purpose of visual analysis is to detect the structure of the document and determine the boundaries of similar regions. Semantic analysis needs to identify specific document elements, such as headings, paragraphs, tables, etc., for these detected areas. PubLayNet~\citep{zhong2019publaynet} is a large-scale document layout analysis dataset. More than 360,000 document images are constructed by automatically parsing PubMed XML files. DocBank~\citep{li-etal-2020-docbank} automatically built an extensible document layout analysis dataset through the correspondence between PDF files and LaTeX files on the arXiv website, and supports both text-based and image-based document layout analysis. IIIT-AR-13K~\citep{mondal2020iiit} also provided 13,000 manually annotated document images for layout analysis.

In Section \ref{sectioncnn}, we introduced the application of the Convolutional Neural Network (CNN) in document layout analysis~\citep{he2015deep,ren2016faster,he2018mask,Liu_2016,yolov3,yang2017learning,8270123}. As the performance requirement for document layout analysis has gradually increased, more research work has made significant improvements with specific detection models.
\cite{Yang2017LearningTE} treat the document semantic structure analysis task as a pixel-by-pixel classification problem. They propose a multimodal neural network that considers both visual and textual information.
\cite{Viana2017FastCD} propose a lightweight model for document layout analysis of mobile and cloud services. This model uses the one-dimensional information of the image for inference, and achieves higher accuracy compared with the model that uses the two-dimensional information.
\cite{chen2017convolutional} introduce a page segmentation method of handwritten historical document images based on CNN.
\cite{oliveira2018dhsegment} propose a multi-task pixel-by-pixel prediction model based on CNN.
\cite{wick2018fully} propose a high-performance Fully Convolutional Neural Network (FCN) for historical document segmentation.
\cite{gruning2019two} propose a two-stage text line detection method for historical documents.
\cite{soto-yoo-2019-visual} incorporate contextual information into the Faster R-CNN model. This model used the local invariance of the article elements to improve the region detection performance.

\paragraph{Table Detection and Recognition}

In document layout analysis, table understanding is an important and challenging subtask. Different from document elements such as headings and paragraphs, the format of the table is usually more variable and the structure is more complex. Therefore, there is a lot of related work carried out around tables, among which the two most important subtasks are table detection and table structure recognition. 1) Table detection refers to determine the boundary of the tables in the document. (2) Table structure recognition refers to extracting the semantic structure of the table, including information about rows, columns, and cells, according to a predefined format.

In recent years, benchmark datasets have emerged for table understanding, including table detection datasets such as Marmot~\citep{fang2012dataset} and UNLV~\citep{10.1145/1815330.1815345}. Meanwhile, the ICDAR conference held several competitions on table detection and recognition, where high-quality table datasets are provided~\citep{Gbel2013ICDAR2T,8978120}. However, these traditional benchmark datasets are relatively small in scale, and it is difficult to unleash the capability of the deep neural networks. Therefore, TableBank~\citep{li-etal-2020-tablebank} uses LaTex and Office Word documents to automatically build a large-scale table understanding dataset. PubTabNet~\citep{zhong2019image} proposes a large-scale table dataset and provides table structure and cell content to assist in table recognition. TNCR~\citep{abdallah2021tncr} provides the label of the table categories while providing the table boundaries.

Many deep learning based object detection models have achieved good results in table detection. Faster R-CNN~\citep{ren2016faster} achieves very good performance by directly applying it to table detection. On this basis, \cite{siddiqui2018decnt} achieve better performance by applying deformable convolution on Faster R-CNN. CascadeTabNet~\citep{prasad2020cascadetabnet} uses the Cascade R-CNN~\citep{cai2018cascade} model to perform table detection and table structure recognition at the same time. TableSense~\citep{dong2019tablesense} significantly improves table detection capabilities by adding cell features and sampling algorithms.

In addition to the above two main subtasks, the understanding of parsed tables has become a new challenge. TAPAS~\citep{herzig2020tapas} introduces pre-training techniques to table comprehension tasks. By introducing an additional positional encoding layer, TAPAS enables the Transformer~\citep{vaswani2017attention} encoder to accept structured table input. After pre-training on a large amount of tabular data, TAPAS significantly surpasses traditional methods in a variety of downstream semantic analysis tasks for tables. Following TAPAS, TUTA~\citep{wang2020structure} introduces a two-dimensional coordinate to represent the hierarchical information of a structured table, and proposes a tree structure based location representation and attention mechanism to show the hierarchical modeling of this structure. Combining different levels of pre-training tasks, TUTA has achieved further performance improvements on multiple downstream datasets.

\subsubsection{Visual Information Extraction}

Visual information extraction refers to the technology of extracting semantic entities and their relationships from a large number of unstructured visually-rich documents. Visual information extraction differs in different document categories and the extracted entities are also different. FUNSD~\citep{jaume2019} is a form understanding dataset that contains 199 forms, where each sample contains key-value pairs of form entities. SROIE~\citep{Huang_2019} is an OCR and information extraction benchmark for receipt understanding, which attracts a lot of attention from the research/industry community. CORD~\citep{park2019cord} is a receipt understanding dataset that contains 8 categories and 54 subcategories of entities. Kleister~\citep{stanislawek2021kleister} is a document understanding dataset for long and complex document entity extraction tasks, including long text documents such as agreements and financial statements. DeepForm~\citep{deepform} is an English dataset for the disclosure form of political advertisements on television. The EATEN dataset~\citep{guo2019eaten} is a dataset for information extraction of Chinese documents. \cite{yu2021pick} further add text box annotations to the 400 subset of EATEN. The EPHOIE~\citep{wang2021towards} dataset is also an information extraction dataset for Chinese document data. XFUND~\citep{xu2021layoutxlm} is a multi-lingual extended version of the FUNSD data set proposed with the LayoutXLM model, which contains visually-rich documents in seven commonly-used languages.

For visually-rich documents, a lot of research models the visual information extraction task as a computer vision problem, and performs information extraction through semantic segmentation or text box detection. Considering that text information also plays an important role in visual information extraction, the typical framework is to treat document images as a pixel grid and add text features to the visual feature map to obtain a better representation. According to the granularity of textual information, these approaches are developed from character-level to word-level and then to context-level. Chargrid~\citep{katti-etal-2018-chargrid} uses a convolution-based encoder-decoder network to fuse text information into images by performing one-hot encoding on characters. VisualWordGrid~\citep{kerroumi2020visualwordgrid} implements Wordgrid~\citep{katti-etal-2018-chargrid} by replacing character-level text information with word-level word2vec features, and fusing visual information to improve the extraction performance. BERTgrid~\citep{denk2019bertgrid} uses BERT to obtain contextual text representation, which further improves the end-to-end accuracy. Based on BERTgrid, ViBERTgrid~\citep{lin2021vibertgrid} fuses the text features from BERT with the image features from the CNN model, thus obtaining better results.

Since textual information still plays an important role in visually-rich documents, a lot of research work takes  information extraction as a special natural language understanding task. \cite{majumder2020representation} generate candidates according to the types of the extracted entities, and has achieved good results in form understanding. TRIE~\citep{zhang2020trie} combines text detection and information extraction, allowing two tasks to promote each other to obtain better information extraction results. \cite{wang2020docstruct} predict the relationship between text fragments through the fusion of three different modalities, and realize the hierarchical extraction for form understanding.

Unstructured visually-rich documents are often composed of multiple adjacent text fragments, so it is also natural to use the Graph Neural Network (GNN) for representation. The text fragments in a document are considered as nodes in the graph, while the relationship between the text fragments can be modeled as edges, so that the entire document can be represented as a graph network. In Section \ref{sectiongnn}, we introduced the representative work of GNN for information extraction in visually-rich documents~\citep{liu-etal-2019-graph}. On this basis, there is more research work based on GNN for visual information extraction. \cite{hwang2020spatial} model the document as a directed graph and extract information from the document through dependency analysis. \cite{riba2019table} use a GNN model to extract tabular information from the invoices. \cite{wei2020robust} use Graph Convolutional Networks (GCN) to model the text layout based on the output of the pre-trained models, which improves information extraction. \cite{cheng2020one} achieves better performance in few-shot learning by representing the document as a graph structure and using a graph-based attention mechanism and a CRF model. The PICK~\citep{yu2021pick} model introduces a graph that can be learned based on nodes to represent documents, and achieved better performance in receipt understanding.

\subsubsection{Document Image Classification}

Document image classification refers to the task of classifying  document images that is essential for business digitalization. RVL-CDIP~\citep{harley2015icdar} is a representative dataset for this task. The dataset contains 400,000 grayscale images in 16 document image categories. Tabacco-3482~\citep{Kumar2014StructuralSF} selects a subset of RVL-CDIP for evaluation, which contains 3,482 grayscale document images.

Document image classification is a special subtask of image classification, thus classification models for natural images can also address the problem of document image classification.
\cite{afzal2015deepdocclassifier} introduce a document image classification method based on CNN for document image classification. To overcome the problem of insufficient samples, they use Alexnet trained with ImageNet as the initialization for model adaptation on document images.
\cite{afzal2017cutting} use GoogLeNet, VGG, ResNet and other successful models from natural images on document images through transfer learning.
Through the adjustment of model parameters and data processing, \cite{tensmeyer2017analysis} use the CNN model that can outperform the previous models without transfer learning from natural images.
\cite{das2018document} propose a new convolutional network based on different image regions for document image classification. This method classifies different regions of the document separately, and finally merges multiple classifiers of different regions to obtain a significant performance improvement in document image classification.
\cite{sarkhel2019deterministic} extract features at different levels by introducing a pyramidal multi-scale structure.
\cite{dauphinee2019modular} obtain the text of the document by performing OCR on the document image, and combine image and text features to further improve the classification performance.

\subsubsection{Document Visual Question Answering}

\begin{figure*}[t]
\centering
    \begin{subfigure}[b]{0.44\textwidth}
        \includegraphics[width=\textwidth]{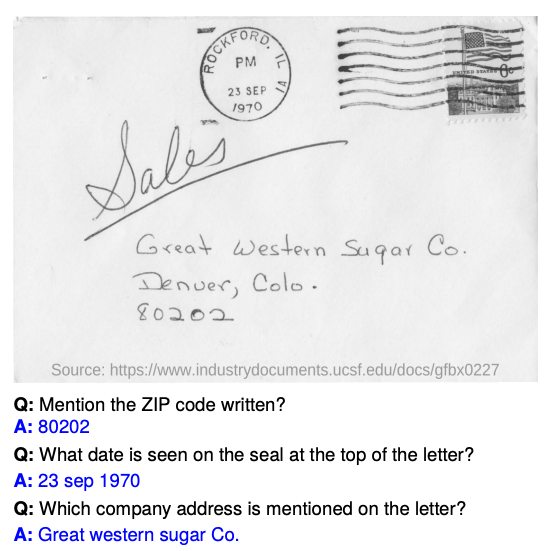}
        \caption{}
        \label{fig:5a}
    \end{subfigure}
    \begin{subfigure}[b]{0.53\textwidth}
        \includegraphics[width=\textwidth]{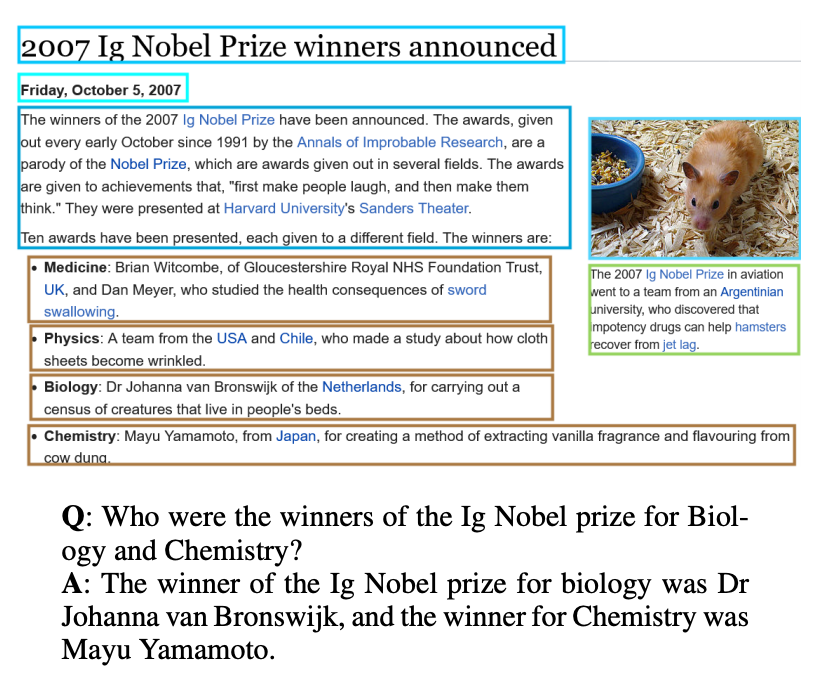}
        \caption{}
        \label{fig:5b}
    \end{subfigure}
    \caption{Examples of Document Visual Question Answering}
    \label{fig:5}
\end{figure*}

Document Visual Question Answering (VQA) is a high-level understanding task for document images. Specifically, given a document image and a related question, the model needs to give the correct answer to the question based on the given image. A specific example is shown in Figure~\ref{fig:5}. VQA for documents first appears in the DocVQA dataset~\citep{mathew2021docvqa}, which contains more than 12,000 documents and corresponding 5,000 questions. Later, InfographicVQA~\citep{mathew2021infographicvqa} is also proposed, which is a VQA benchmark for infographic images in the documents. As the answers in DocVQA are relatively short and topics are not diverse, some researchers also proposed the VisualMRC~\citep{tanaka2021visualmrc} dataset for the document VQA task, which includes long answers with diverse topics.

Different from the traditional VQA task, textual information in document VQA plays a key role in this task, so existing representative methods all take the texts obtained by OCR of document images as the inputs. After the document text is obtained, the VQA task is modeled as different problems according to the characteristics of different datasets. For the DocVQA data, most of the answers to questions exist as text fragments in the document text, so mainstream methods have modeled it as the Machine Reading Comprehension (MRC) problem. By providing the model with visual features and document texts, the model extracts text fragments from the given document according to the question as the corresponding answer. For the VisualMRC dataset, the answer to the question usually does not literally appear in the document text fragment and a longer abstract answer is required. Therefore, a feasible method is to use a text generation approach to generate answers to the questions.

\subsection{General-purpose Multimodal Pre-training}

Although the above methods achieve good performance on document understanding tasks, these methods usually have two limitations: 1) The models often rely on limited labeled data, while neglecting a large amount of knowledge in unlabeled data. On one hand, for document understanding tasks such as information extraction, human annotation of data is expensive and time-consuming. On the other hand, due to the extensive use of visually-rich documents in the real world, there are a large number of unlabeled documents, and these large amounts of unlabeled data can be leveraged for self-supervised pre-training. 2) Visually-rich documents not only contain a lot of text information, but also rich layout and visual information. Existing models for specific tasks usually only use pre-trained CV models or NLP models to obtain the knowledge from the corresponding modality due to the limitation of data, and most of the work only uses information from a single modality or simple combination of features rather than the deep fusion. The success of Transformer~\citep{vaswani2017attention} in transfer learning proves the importance of deep contextualization for sequence modeling for both NLP and CV problems. Therefore, it is obvious to jointly learn different modalities such as text, layout and visual information in a single framework.

Visually-rich documents mainly involve three modalities: text, layout, and visual information, and these modalities have a natural alignment in visually-rich documents. Therefore, it is vital to model document representations and achieve cross-modal alignment through pre-training. The LayoutLM~\citep{10.1145/3394486.3403172} and the subsequent LayoutLMv2~\citep{xu-etal-2021-layoutlmv2} model are proposed as the pioneer work in this research area. In Section~\ref{sectiontransformer}, we introduced LayoutLM, a general pre-trained model for Document AI. Through joint pre-training of text and layout, LayoutLM has achieved significant improvement in a variety of document understanding tasks. On this basis, there is a lot of follow-up research work to improve this framework. LayoutLM does not introduce document visual information in the pre-training process, so the accuracy is not satisfactory on tasks that require strong visual perception such as DocVQA. In response to this problem, LayoutLMv2~\citep{xu-etal-2021-layoutlmv2} integrates visual information into the pre-training process, which greatly improves the visual understanding capability. Specifically, LayoutLMv2 introduces a spatial-aware self-attention mechanism, and uses visual features as part of the input sequence. For the pre-training objectives, LayoutLMv2 proposes ``Text-Image Alignment'' and ``Text-Image Matching'' tasks in addition to Masked Visual-Language Modeling. Through improvements in these two aspects, the model capability to perceive visual information is substantially improved, and it significantly outperforms strong baselines in a variety of downstream Document AI tasks.

Visually-rich documents can be generally divided into two categories. The first one is the fixed-layout documents such as scanned document images and digital-born PDF files, where the layout and style information is pre-rendered and independent of software, hardware, or operating system. This property makes existing layout-based pre-training approaches easily applicable to document understanding tasks. While, the second category is the markup language based documents such as HTML/XML, where the layout and style information needs to be interactively and dynamically rendered for visualization depending on the software, hardware, or operating system. For markup language based documents, the 2D layout information does not exist in an explicit format but usually needs to be dynamically rendered for different devices, e.g. mobile/tablet/desktop, which makes current layout-based pre-trained models difficult to apply. To this end, MarkupLM~\citep{li2021markuplm} is proposed to jointly pre-train text and markup language in a single framework for markup-based VrDU tasks.  Distinct from fixed-layout documents, markup-based documents provide another viewpoint for the document representation learning through markup structures because the 2D position information and document image information cannot be used straightforwardly during the pre-training. Instead, MarkupLM takes advantage of the tree-based markup structures to model the relationship among different units within the document.

\paragraph{Position Information}
After LayoutLM, much research work has made improvements based on this model framework. One of the main directions is to improve the way of position embeddings. Some work has changed the position encoding represented by embeddings to the sinusoidal functions, such as BROS~\citep{hong2020bros} and StructuralLM~\citep{li2021structurallm}. BROS~\citep{hong2020bros} uses the sinusoidal function for the absolute position encoding, and at the same time introduced the relative position information of the text through the sinusoidal function in the self-attention mechanism, which improves the model's ability to perceive spatial position. StructuralLM~\citep{li2021structurallm} shares the same position information in the text block in the absolute position representation, which helps the model understand the text information in the same entity, thereby further improving information extraction.

\paragraph{Visual Information}
In addition, some research work has made further improvements to optimize and strengthen the vision models. LAMPRET~\citep{wu2021lampret} provides the model with more visual information to model web documents such as font size, illustrations, etc. which helps to understand rich web data. SelfDoc~\citep{li2021selfdoc} adopts a two-stream structure. For a given visually-rich document image, a pre-trained document entity detection model is first used to identify all semantic units in the document through object detection, then OCR is used to recognize the textual information. For the identified image regions and text sequences, the model uses Sentence-BERT~\citep{reimers-gurevych-2019-sentence} and Faster-RCNN~\citep{ren2016faster} to extract features and encode them as feature vectors. A cross-modal encoder is used for encoding the whole image with a multi-modal representation to serve downstream tasks. DocFormer~\citep{appalaraju2021docformer} adopts a discrete multi-modal structure and uses position information on each layer to combine text and visual modalities for the self-attention. DocFormer uses ResNet~\citep{he2015deep} to encode image information to obtain higher resolution image features, and at the same time encodes text information into text embeddings. The position information is added to the image and text information separately and passed to the Transformer layer separately. Under this mechanism, high-resolution image information was obtained while the input sequence was shortened. Meanwhile, different modalities are aligned through position information so that the model could better learn the cross-modal relationship of visually-rich documents.

\paragraph{Pre-training Tasks}
Moreover, some pre-trained models have designed richer pre-training tasks for different modalities.
For example, in addition to the Masked Visual-Language Modeling (MVLM), BROS~\citep{hong2020bros} proposes an area-masked language model, which masks all text blocks in a randomly selected area. It can be interpreted as extending the interval mask operation for one-dimensional text in SpanBERT~\citep{Joshi2020SpanBERTIP} to an interval mask for text blocks in a two-dimensional space. Specifically, the operation consists of the following four steps: (1) randomly selecting a text block, (2) determining a final area by expanding the area of the text block, (3) determining the text block belonging to the area, (4) masking all the text within block and recover.
LAMPRET~\citep{wu2021lampret} additionally introduces the ordering of web page entities, which allows the model to learn the spatial position by predicting the order of the entity arrangement. At the same time, the model also uses image matching pre-training by removing images in the webpage and matching through retrieval. This also improves the model's ability to understand the semantics of multimodal information.
The ``Cell Position Classification'' task proposed by StructuralLM~\citep{li2021structurallm} models the relative spatial position of the text block in the document. Given a set of scanned files, this task aims to predict the location of text blocks in the file. First, a visually-rich document is divided into N regions of the same size. Then, the model calculates the area to which the text block belongs through the two-dimensional position of the center of the text block.
SelfDoc~\citep{li2021selfdoc} and DocFormer~\citep{appalaraju2021docformer} also introduce new pre-training tasks along with the improvements of image inputs. SelfDoc masks and predicts the image features to better learn the visual information. DocFormer introduces a decoder to reconstruct image information. In this case, the task is similar to the image reconstruction of an autoencoder, but it contains multimodal features such as texts and positions. With the help of joint image and text pre-training, image reconstruction requires the deep fusion of texts and images, which strengthens the interaction between different modalities.

\paragraph{Initialization}
Regarding model initialization, some approaches used the existing powerful pre-trained language models to further improve their performance, while also expanding the capabilities of the pre-trained models. For example, LAMBERT~\citep{garncarek2020lambert} achieves better performance by using RoBERTa~\citep{Liu2019RoBERTaAR} as the pre-training initialization. In addition to language understanding, some models focus on extending the language generation capabilities of the models. A common practice is to use the encoder-decoder models for initialization. TILT~\citep{powalski2021going} introduces the layout encoding layer into the pre-trained T5~\citep{2020t5} model and combined document data for pre-training, so that the model can handle the generation tasks in Document AI. LayoutT5 and LayoutBART~\citep{tanaka2021visualmrc} introduce text position encoding on top of T5~\citep{2020t5} and BART~\citep{lewis-etal-2020-bart} models in the fine-tuning stage for document VQA to help the model understand questions and generate answers better.

\paragraph{Multilingual}
Although these models have been successfully applied on English documents, document understanding tasks are also important for non-English speaking worlds. LayoutXLM~\citep{xu2021layoutxlm} is the first research work that carries out multilingual pre-training on visually-rich documents for other languages. Based on the model structure of LayoutLMv2, LayoutXLM expands the language support of LayoutLM by using 53 languages for pre-training. Compared with the cross-language models for plain text, LayoutXLM has obvious advantages in the language expansion capability for visually-rich documents, which proves that cross-lingual pre-training not only works on pure NLP tasks, but also effective for cross-lingual Document AI tasks.

\section{Conclusion and Future Work}

Automated information processing is the foundation and prerequisite for digital transformation. Nowadays, there are increasingly higher requirements for processing power, processing speed, and processing accuracy. Taking the business field as an example, electronic business documents cover a large amount of complicated information such as purchase receipts, industry reports, business emails, sales contracts, employment agreements, commercial invoices, and personal resumes. The Robotic Process Automation (RPA) industry has been created in this background, using AI technology to help a large number of people free from complicated electronic document processing tasks, meanwhile improving productivity through a series of supporting automation tools. One of the key cores of RPA is the Document AI technique. In the past 30 years, document analysis has mainly gone through three stages, from the early-stage heuristic rules, to the statistical machine learning, and recently deep learning methods, which greatly advances analysis performance and accuracy. At the same time, we have also observed that the large-scale self-supervised general document-level pre-trained model represented by LayoutLM has also received more and more attention and usage, and has gradually become the basic unit for building more complex algorithms. There are also quite a few follow-up research work that emerged recently, which accelerates development of the Document AI.

For future research, in addition to the multi-page/cross-page problems, uneven quality of training data, weak multi-task relevance, and few-shot and zero-shot learning, we also need to pay special attention to the relationship between OCR with Document AI tasks, since the input of Document AI applications usually comes from automatic OCR models. The accuracy of text recognition often has a great impact on downstream tasks. In addition, how to combine Document AI technology with existing human knowledge especially manual document processing skills, is an interesting research topic worth exploring in the future. 


\bibliography{iclr2021_conference}
\bibliographystyle{iclr2021_conference}


\end{document}